\documentclass{article}
\usepackage[preprint]{neurips_2023}
\usepackage[hyphens]{url}
\usepackage{graphicx}
\usepackage{caption}
\usepackage{newfloat}
\DeclareFloatingEnvironment[fileext=loa,placement={tbp},name=Algorithm]{algo}
\usepackage{booktabs}
\usepackage{amsmath}
\usepackage{amssymb}
\usepackage{array}
\usepackage{pifont}
\usepackage{algpseudocode}
\algrenewcommand\algorithmicrequire{\textbf{Input:}}
\algrenewcommand\algorithmicensure{\textbf{Output:}}
\usepackage[hidelinks]{hyperref}

\makeatletter
\@addtoreset{subparagraph}{subsection}

\makeatother
\newcolumntype{L}[1]{>{\raggedright\arraybackslash}p{#1}}

\title{Agentic Empirical Asset Pricing:\\Methodological Foundations\thanks{\textbf{AI usage disclosure.} Generative AI tools were used as an assistant in this project's process---for part of the supporting code development and for portions of this manuscript's preparation and editing. The human authors reviewed all AI-assisted artifacts and take full responsibility for the paper's content (full detail in the AI Acknowledgement, Appendix~\S\ref{app:ai-ack}).}}

\author{
Yingjian Pan\\
{\small Advanced Financial Technologies Laboratory}\\
{\small Management Science \& Engineering}\\
{\small Stanford University}\\
{\small \texttt{pyj@stanford.edu}}
\And
Xiaowei Ding\\
{\small School of Information Management}\\
{\small Nanjing University}\\
{\small \texttt{dingxiaowei@nju.edu.cn}}
\AND
Kay Giesecke\\
{\small Artificial Intelligence \& Quantitative Finance}\\
{\small Hasso Plattner Institute}\\
{\small \texttt{kay.giesecke@hpi.de}}
}
\date{}

\begin{document}
\maketitle

\begin{center}
This draft: July, 2026
\end{center}
\vspace{3mm}

\begin{abstract}
Recent advances in LLM agents enable a new paradigm for asset pricing, which we call Agentic Empirical Asset Pricing (AEAP): systems that autonomously conduct the scientific discovery process itself. We define AEAP and identify its core building blocks. Existing evaluation practices backtest only the outputs---factors or trades---not the autonomous discovery system that produced them. We focus on factor discovery, contributing a reference architecture, a rigorous evaluation standard for discovered factors, and a method for out-of-sample backtesting the discovery system. As a concrete instance of that architecture, we evaluate SEADS against five re-implemented baselines on two US equity panels using this standard: no single metric ranks the systems consistently, motivating evaluation on multiple axes at once. A separate rolling re-execution then asks the complementary question of whether the discovery process itself, not one static output, is reliable. We also report negative findings and limitations that surface further evaluation pitfalls for future AEAP systems.
\end{abstract}

\section{Introduction}
\label{sec:intro}

Empirical asset-pricing research has traditionally relied on humans to design and execute the discovery process, whether through theory-guided construction or data-driven machine learning \citep{gukellyxiu2020,kozaknagelsantosh2020}. LLM agents that autonomously generate hypotheses, synthesize them into executable code, and statistically validate the outcome change this division of labor---we call the resulting paradigm \textbf{Agentic Empirical Asset Pricing (AEAP)} (\S\ref{sec:aap}).

The paradigm is not hypothetical: in the last two years, at least a dozen systems have implemented a hypothesize--code--backtest--refine loop for formulaic factor mining \citep{alphaagent2025,alphaforge2025,rdagentq2025,quantaalpha2026,beyondprompting2026,alphajungle2025}, multi-agent teams reach trading decisions through debate or structured pipelines \citep{tradingagents2024,multiagentport2026,sharp2026}, and LLM agents are wired into structural pricing pipelines built around an explanatory asset-pricing model \citep{koijenlevy2026,chengchin2024aapm}. The resemblance across these is real but the vocabulary is not shared: each names its own architecture, states its own notion of validity, and is evaluated against its own idiosyncratic metric.

That last point is what this paper is built around: a shared vocabulary's absence is not just inconvenient, it hides a specific failure mode. Every AEAP system we know of is evaluated the way a classical, human-discovered factor is: hold out a test window, backtest the \emph{output}---the factor or trade---and report its performance. This is necessary but not sufficient. An autonomous discovery system is itself a research process with its own way of getting lucky: a gate admitting only a small fraction of one run's candidates can be well-calibrated, can be leaking information through an unnoticed bug, or can simply have drawn a favorable in-sample window---nothing about that run's own reported Sharpe distinguishes the three. None of this is visible from backtesting the output alone---it requires backtesting the \emph{system that produced it}, re-run under the information constraints a real deployment would face. To our knowledge, no existing AEAP evaluation does this. The problem is not specific to finance: any autonomous discovery agent that revises its own hypotheses over time faces the same question, and asset pricing simply gives it an unusually clean testbed---clear temporal ordering, measurable future outcomes, and severe adaptive-overfitting risk.

That gap is a foundations problem before it is a better-system problem. Without a shared definition of what counts as an AEAP system, an account of the objectives it pursues, and a standard for scoring the discovery process itself rather than only its output, results across systems cannot be compared, replicated, or built upon---each new entrant restarts the vocabulary from scratch. This paper supplies those foundations: broadly, for the paradigm as a whole (\S\ref{sec:aap}); in full technical detail, for factor discovery---EAP's most central objective, hence most agentic work's focus (\S\ref{sec:method}).

\paragraph{Contributions.} We make three.

\textbf{(C1) A definition of AEAP and its building blocks.} We define AEAP precisely enough to distinguish it from both classical ML asset pricing and LLM copilots that leave a human in the loop, and identify five building blocks common to any such system (\S\ref{sec:aap}).

\textbf{(C2) For factor discovery: an evaluation methodology.} We factor out of five re-implemented systems (\S\ref{sec:related}) a reference architecture, then consolidate a multi-part evaluation standard around three jointly-necessary properties of a discovered pool---productivity, performance, and novelty, no one meaningful without the other two---for scoring what a discovery system reports. We further propose re-executing the \emph{entire} discovery loop at a sequence of historical decision dates: measuring whether a discovery process adapts over time rather than getting lucky once, a distinction no static backtest can draw (\S\ref{sec:method}).

\textbf{(C3) SEADS: a concrete agentic factor-discovery system.} We build \textbf{SEADS} (Stanford Engine for Agentic Discovery of Signals, pronounced ``seeds''---an echo of the Planner's own seed records, \S\ref{sec:seads-gen}) as an instance of (C2)'s reference architecture and apply (C2)'s standard to it alongside five re-implemented baselines \citep{alphaagent2025,alphaforge2025,rdagentq2025,quantaalpha2026,beyondprompting2026} on two comprehensive US equity panels: the US subset of JKP's global-characteristics panel \citep{jensenkellypedersen2023}, and a CRSP/Compustat panel of primitive features novelty-checked against the 45-characteristic reference panel of \citet{bryzgalova2025missing}. No single metric ranks all six systems consistently, which is itself the point of (C2)'s standard (\S\ref{sec:main}). We also surface negative findings and limitations that are pitfalls for any discovery system, not only ours (\S\ref{sec:limits}).

\section{Agentic Empirical Asset Pricing}
\label{sec:aap}

\subsection{Definition and Core Building Blocks}
\label{sec:aap-def}

Empirical Asset Pricing (EAP) is the branch of financial economics seeking scientific understanding of asset prices through data. Its objectives recur across the field, e.g.\ testing asset-pricing theories (CAPM, ICAPM, consumption-based models), discovering and evaluating risk factors \citep{jensenkellypedersen2023}, explaining the cross-section of expected returns, forecasting returns, studying anomalies, evaluating portfolio strategies, and estimating risk premia. These objectives were traditionally pursued through parametric econometrics; over the last decade, a second wave used data-driven machine learning \citep{gukellyxiu2020,kozaknagelsantosh2020}. We propose a third wave: \textbf{Agentic Empirical Asset Pricing (AEAP)}, in which autonomous LLM-based agents pursue these objectives with minimal human intervention---not displacing econometrics and ML but calling on them within an agent's own loop. We propose not a new objective, but a new researcher.

Formally, we define an \textbf{AEAP system} as one in which LLM-based agents autonomously execute at least one full cycle of the scientific-discovery process---hypothesis, formalization, evaluation---over asset-pricing data, without a human performing any of those three steps. This has two consequences for what counts. A machine-learning return predictor \citep{gukellyxiu2020,kozaknagelsantosh2020} is not an AEAP system: the researcher, not the model, designs the hypothesis space and decides what counts as validation. A human proposing an idea for an LLM copilot to formalize and backtest \citep{alphagpt2023} is likewise not fully autonomous. What qualifies is a system whose loop---propose, formalize, check---closes without a human decision, even if a human designed it and reviews its output; a single cycle already qualifies, and persisting state across cycles (B5) is common, not required.

Reading across the systems we survey (\S\ref{sec:related}, Appendix~\ref{app:landscape}), five building blocks recur regardless of output: \textbf{(B1) hypothesis generation}---an LLM proposes a candidate mechanism (a factor's rationale, a trade thesis, an anomaly's explanatory channel), not a bare parameter draw; \textbf{(B2) formalization} into a checkable artifact---code for a factor, a discrete action for a trade, a fitted model for a structural claim; \textbf{(B3) execution or estimation} against historical or live data---simulated for a factor or trade, statistically fit for a structural model; \textbf{(B4) an evaluation gate}, statistical or LLM-judged, deciding what survives or, for debate-structured systems, what the group commits to; and \textbf{(B5) memory and iteration}: persistent state---including the discovered set and its reference pool, not only model weights---that lets the system improve across cycles instead of restarting cold. B1--B2 generate; B4 judges; keeping them separate makes B4's verdict evidence, not the generator's own opinion of itself, whether B4 is a statistical test or an independent LLM critic. These map onto standard LLM-agent components from the broader agentic-AI literature \citep{wangautonomousagents2024}: B1--B2 to planning and action-formulation, B3 to tool use, B4 to a critic model, B5 to memory---grounding AEAP in agentic AI generally, not only in finance. B3 and B4 are stated here in their most literal, factor-discovery-native form.

\subsection{Examples}
\label{sec:related}

AEAP is not specific to one EAP objective: discovering factors is scored as a cross-sectional signal, evaluating a trading policy as a trajectory of decisions, testing a structural model by estimation. The systems below span these objectives; not all are AEAP systems.

\paragraph{Factor discovery.} \textbf{AlphaAgent} \citep{alphaagent2025} gates on AST-originality and hypothesis-factor consistency; \textbf{RD-Agent(Q)} \citep{rdagentq2025} co-optimizes factors and models via bandit-scheduled feedback and a backtest-metric admission unit; \textbf{QuantaAlpha} \citep{quantaalpha2026} evolves mining trajectories under LLM-judged semantic consistency plus algorithmic complexity/redundancy constraints; \textbf{Beyond Prompting} \citep{beyondprompting2026} is, on its own account, this group's rigor frontier, importing the \citet{harveyliuzhu2016} $t>3.0$ hurdle; and \textbf{AlphaForge} \citep{alphaforge2025}---non-LLM, and not itself an AEAP system by \S\ref{sec:aap-def}'s definition, but included as the field's most-cited non-agentic combination baseline---mines via a generative-predictive network and combines survivors dynamically. None combines more than two independent statistical criteria into one decision, and none re-executes its own discovery loop at more than one historical decision date (Table~\ref{tab:comparison}, Appendix~\ref{app:refarch-audit}). AlphaForge's mining stage instead searches combinatorially, filtering ex post: more candidates only raise the bar a genuine discovery must clear.

\paragraph{Other systems.} \citet{tradingagents2024,multiagentport2026,sharp2026} reach trading decisions through multi-agent debate or structured pipelines, and \citet{koijenlevy2026,chengchin2024aapm} estimate an explanatory asset-pricing model end-to-end---genuine AEAP systems (\S\ref{sec:aap-def}), pursuing a different objective, not directly comparable to ours. Further removed, and not meeting the AEAP standard themselves: non-agentic predecessors \citep{alphagen2023,alphaevolve2021} lack the agentic loop, and general-purpose ``AI-scientist'' engines \citep{aiscientist2024,aiscientistv2_2025,alphaevolve2025deepmind} lack the finance-specific objective, though both are architecturally foundational; Appendix~\ref{app:landscape} extends this survey substantially. We further situate this work within the ML asset-pricing lineage \citep{gukellyxiu2020,kozaknagelsantosh2020} and the replication debate \citep{jensenkellypedersen2023}.

\subsection{Evaluation Principles}
\label{sec:principles}

Having surveyed these systems, we now ask how any of them---or SEADS---should be evaluated. Classical backtesting reduces to one discipline: use only information available at each decision date. When the researcher is itself a model, this discipline binds two nested loops, not one. An \textbf{inner loop} is the discovery cycle itself, hypothesis through evaluation; an \textbf{outer loop} re-runs it across historical decision dates rather than scoring output on just one split. A methodologically sound AEAP evaluation needs three principles: \textbf{(P1) point-in-time data}---no feature observable only after decision date $t$ enters the inner loop; \textbf{(P2) point-in-time process}---the outer loop actually exists: the inner loop is re-executable and scoreable at each historical date, not just run once; \textbf{(P3) out-of-sample-only results}---headline numbers come from a split neither loop ever touched. Classical backtests satisfy P1 and P3 by construction, and never needed an outer loop either: with a human, not an autonomous inner loop, doing the discovering, there was no comparably systematic process to re-run. \textbf{P2 is the property no existing AEAP system we survey satisfies} (Table~\ref{tab:comparison}, Appendix~\ref{app:refarch-audit})---an AEAP system's own \emph{reliability} cannot be assessed without it (\S\ref{sec:method-backtest}). We omit a fourth, backbone-vintage principle; \S\ref{sec:leakage} argues its residual risk is task-dependent---\emph{soft} for ours.

We focus the rest of this paper on factor discovery, the objective with the most existing systems and the best-defined statistical target for \S\ref{sec:method}'s backtesting-the-system question.

\section{Factor Discovery}
\label{sec:method}

\S\ref{sec:principles}'s P1--P3 principles are not specific to factor discovery; they generalize to any AEAP objective. We narrow to factor discovery here because it is the objective most tractable to formalize fully, end to end.

\subsection{Problem Formulation}
\label{sec:method-def}
We work on a panel of asset-periods $(i,t)$ with features $x_{i,t}\in\mathbb{R}^p$ observable at the close of period $t$ and next-period excess return $r_{i,t}$ over $(t,t{+}1]$; our own panels (\S\ref{sec:setup}) take the period to be a month, but nothing in the formulation requires it. A candidate factor is a scoring function $f_\theta: x_{i,t}\mapsto s_{i,t}$, typically ending in a within-period percentile rank, where $\theta$ denotes a symbolic (Python) recipe rather than a learned parameter vector; its period-$t$ information coefficient is $\rho_t=\mathrm{Spearman}(s_{\cdot,t},r_{\cdot,t})$ across stocks---the quantity \S\ref{sec:seads-gate}'s Rank-IC check is built on.

The discovery target is a factor set $\mathcal{G}=\{f_{\theta_1},\ldots,f_{\theta_q}\}$ that is simultaneously out-of-sample valid and non-redundant; no single metric captures both (\S\ref{sec:method-standard}). Non-redundancy is with respect to an explicit, disclosed reference set $\mathcal{Z}$, chosen per panel rather than drawn from a single run's own candidates, which would be circular (\S\ref{sec:setup} gives each panel's specific choice); $f_\theta$ is admissible only if $|\mathrm{corr}(s_{\cdot,t},z_{\cdot,t})|$ stays below a threshold for every $z\in\mathcal{Z}$ (\S\ref{sec:seads-gate}). $\mathcal{G}$ and $\mathcal{Z}$ are both fixed here at one decision date; \S\ref{sec:method-backtest} asks what changes once both are allowed to evolve across a rolling history instead.

\subsection{A Factor-Discovery Reference Architecture}
\label{sec:refarch}
\begin{figure*}[t]
\centering
\includegraphics[width=0.98\textwidth]{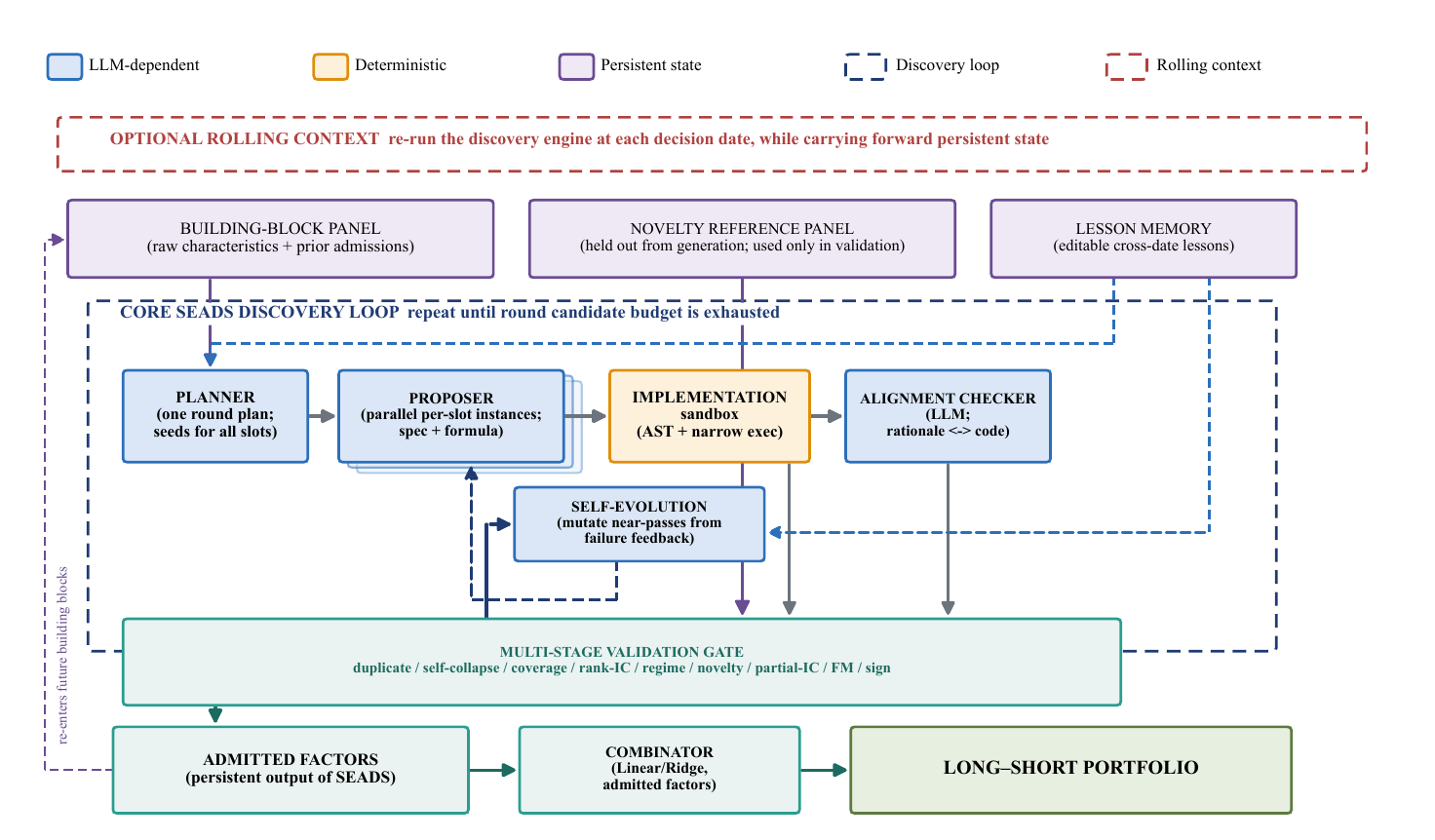}
\caption{Diagram for SEADS's architecture. One Planner call plans an entire round (one seed per slot); each seed is then expanded by its own Proposer instance (stacked boxes). Near-passes are mutated by self-evolution and re-run through the identical sequence. The dashed navy loop is the discovery-round unit every system in Table~\ref{tab:comparison} has; the rolling-context banner above it is SEADS-specific (none of the five baselines re-executes this loop, Table~\ref{tab:refarch-audit}), so it is a light annotation, not a nested box. Figure~\ref{fig:wf-timeline} (Appendix~\ref{app:wf-protocol}) details the repropose/refit mechanics; Algorithm~\ref{alg:seads} (Appendix~\ref{app:algorithm}) gives it as pseudocode. The gate box lists its nine checks; \S\ref{sec:seads-gate} gives the full order and thresholds.}
\label{fig:arch}
\end{figure*}
Table~\ref{tab:comparison}'s (Appendix~\ref{app:refarch-audit}) five baselines share no common vocabulary. Reading across all five re-implementations (four agentic, one not; \S\ref{sec:related}), each instantiates \S\ref{sec:aap-def}'s building blocks as a \textbf{generator} (B1--B2), a \textbf{sandboxed executor} against point-in-time data (B3), a \textbf{validation gate} (B4), and, in the more developed systems, a \textbf{persistence mechanism} accumulating admitted candidates into a library (B5). A well-designed B3 is ideally schema-agnostic, so the same executor and gate generalize across feature panels without redesign---the property \S\ref{sec:seads-sanity} details for SEADS's own implementation. This is also why we exclude \textbf{CogAlpha} \citep{cogalpha2025} from direct comparison: its executor is built around daily OHLCV series, not a schema-agnostic characteristic panel, so porting it would mean reimplementing its core mechanism, not just retargeting its gate. Figure~\ref{fig:arch} diagrams SEADS's own architecture, one concrete instantiation of this pattern; no specific module is mandated---any can be swapped for another implementation while the system remains an instance of it---but Table~\ref{tab:comparison} shows every baseline instantiates only a strict subset (\S\ref{sec:related}).

\subsection{SEADS: An Instantiation}
\label{sec:seads}
SEADS instantiates \S\ref{sec:refarch}'s factor-discovery architecture concretely (Figure~\ref{fig:arch}); it is the system we build and evaluate, and Algorithm~\ref{alg:seads} (Appendix~\ref{app:algorithm}) states its full discovery loop as pseudocode. Data construction for both panels it runs on is in \S\ref{sec:setup} and Appendix~\ref{app:data}.

\subsubsection{Generation: Planner and Proposer}
\label{sec:seads-gen}
Each round, a single Planner call emits one \emph{seed} per candidate slot---a structured record (explore-or-exploit type, theme, variables, mechanism, complexity tier) rather than a finished formula. Seeds split between \emph{exploit} (a bounded variation of a prior successful recipe) and \emph{explore} (a hypothesis from a different signal family, default 70\% of slots); two deterministic backstops correct observed Planner failure modes (Appendix~\ref{app:internals}). Planner and Proposer are deliberately separate roles, not one call issuing finished proposals directly: the Planner controls breadth---how slots split across themes and complexity tiers---while each seed's own Proposer instance controls depth. The Proposer draws on a narrower context scoped to prior attempts sharing at least one underlying variable, not the Planner's full history (Appendix~\ref{app:internals}), independently expanding its seed into a complete specification and formula, which then proceeds to sanity checks (\S\ref{sec:seads-sanity}).

\subsubsection{Sanity Checks}
\label{sec:seads-sanity}
A deterministic structural pre-check first rejects malformed specs before any code executes (Appendix~\ref{app:internals}). Point-in-time data (P1, \S\ref{sec:principles}) is enforced next: the Implementation Agent runs a surviving spec's formula in a sandboxed namespace restricted to exactly the columns it statically reads, checks it against a look-ahead firewall, and attempts up to two LLM-assisted repairs on failure. The firewall is a deterministic AST walk enforcing seven rule classes plus a taxonomy whitelist (Appendix~\ref{app:gatefirewall}); nothing in the sandbox, prompts, or gate hardcodes column names, so the identical pipeline runs unmodified over JKP's $\sim$400 characteristics or CRSP/Compustat's 87 variables alike (\S\ref{sec:setup}). An Alignment Checker---an LLM call---separately verifies the formula implements what its rationale describes via a required line-by-line trace, rejecting only at high confidence.

\subsubsection{Validation Gate}
\label{sec:seads-gate}
Once a candidate clears the sanity checks above, it faces the statistical validation gate, admission requiring all checks to pass, in order. Two structural checks come first: a duplicate fingerprint, and a self-collapse check rejecting a candidate whose final formula algebraically collapses to a plain rank of one of its own declared inputs. Three core statistical checks follow: coverage, Rank-IC $t$-stat, and out-of-decade regime-consistency. A final robustness trio closes the gate---novelty against the reference set $\mathcal{Z}$ (\S\ref{sec:method-def}), partial-IC against the top-$K$ most-correlated controls, and Fama--MacBeth \citep{famamacbeth1973} $t$ against the same controls---with sign agreement required among Rank-IC, partial-IC, and Fama--MacBeth beta (full order and thresholds in Appendix~\ref{app:gatefirewall}). The out-of-decade split deliberately tests cross-regime generalization, not same-period consistency. Each threshold is set from discipline convention, an established statistical procedure (the Rank-IC bar's Bonferroni correction), or a simple ex-ante default, fixed before any experimental run and never adjusted afterward. No baseline combines this many independent checks into one admission decision (\S\ref{sec:related}).

\subsubsection{Self-Evolution}
\label{sec:seads-evolve}
Self-evolution mutates a near-pass---one that cleared a minimum progress floor or failed at a specific gate stage---through the identical gate again, for a bounded number of rounds, stopping early once a mutation clears the Rank-IC bar or shows no further improvement across attempts; candidates falling far short of any bar are discarded rather than mutated. Feedback names the exact failure stage rather than a scalar fitness, and a progress-weighted score rewards reaching a later stage over clearing an earlier one by a larger margin (Appendix~\ref{app:internals}).

\subsubsection{Persistent State and Combination}
\label{sec:seads-state}
Three memory tiers persist across rounds: baseline characteristic statistics the Planner uses for new seeds; an append-only log of every candidate's outcome---admission, failure mode, or mutation---consulted to avoid repeating dead ends; and capped natural-language lessons from a reflector, feeding seeds. Admitted factors accumulate in a persistent library; a combinator fits a Ridge regression ($\alpha=1$) on the library's rank-normalized values to rank stocks monthly into a long-short portfolio---the deployed portfolio construction Table~\ref{tab:main}'s performance column scores. This is simpler by design than alternatives like XGBoost or LightGBM: an ex-ante, reproducibility-motivated default, fixed pre-run (Appendix~\ref{app:internals}) and distinct from \S\ref{sec:wf}'s joint regression, a separate external incremental-value check.

\subsection{Evaluating Discovered Factors}
\label{sec:method-standard}
However a system's output was produced, how should it be scored? A pool of discovered factors earns its place in a fund only if it clears three bars jointly: \textbf{productivity}, \textbf{performance}, and \textbf{novelty}---no two substitute for the third. Performance without novelty rewards redundant repackaging of what a fund already trades (\S\ref{sec:main}'s account of Beyond Prompting's larger, markedly less-novel admitted set). Performance without a productivity expectation rewards finding one good signal by any means---a bar so low that AlphaForge's non-agentic search, with only 5 and 1 total admissions on our two panels (Table~\ref{tab:main}), already clears it, and an agentic system that cannot out-produce that has no case for its own expense. Novelty without performance rewards noise, novelty's cheapest source. \S\ref{sec:method-def}'s two discovery desiderata defined what makes any one factor good: out-of-sample validity and non-redundancy. Scoring a \emph{pool} carries both forward---validity becomes \emph{performance}, non-redundancy becomes \emph{novelty}---and adds \emph{productivity} as the genuinely new question a set, not a single factor, raises. Each of the three admits more than one reasonable measurement---performance alone could be scored by mean or median per-factor Sharpe, by information coefficient, or by post-admission survival, to name a few---so the metrics below are our chosen operationalization, not the only defensible one.

We score what a factor-discovery system reports---SEADS and the five baselines alike---on four metrics (Table~\ref{tab:main}): \emph{productivity}, total admissions out of the shared 300-candidate budget---a function of both native gate strictness and discovery quality, which is why we never score it alone; \emph{performance}, mean per-factor out-of-sample (OOS) Sharpe across the admitted set; and \emph{novelty}, each admitted factor's maximum correlation against the union of the system's own other admits and the panel's reference set, averaged across the admitted set rather than worst-cased so one highly-correlated factor cannot sink an otherwise-diversified system's score---one metric per jointly-necessary property. A fourth check tests performance and novelty jointly: a \emph{spanning test}, a Ridge regression of the system's admitted factors against the reference set: the two resulting OOS long-short return series are regressed against each other, and we report the $t$-statistic on the alpha. Larger positive values are read as suggestive, not confirmatory, evidence of value added beyond the reference set. We score SEADS and all five baselines identically.

\subsection{Backtesting the Discovery System}
\label{sec:method-backtest}
\S\ref{sec:method-standard} scores whatever a system reports, however its output was produced---but productivity, performance, and novelty are all properties of one decision date's output, and factor discovery is not, in practice, a one-off exercise: markets shift, and a deployed system must keep finding and refreshing signals as they do. A fourth dimension---call it \emph{adaptivity}---asks whether the discovery \emph{process itself} both holds up across changing conditions and improves by carrying forward what it has already learned, neither of which a static snapshot, however many axes it scores, can show by construction. SEADS's own gate already tests a static proxy for the first half---requiring a candidate's Rank-IC to hold out-of-decade, not only within-period (\S\ref{sec:seads-gate})---but one regime-robust admission is still one decision at one point in time, saying nothing about whether the system uses what it already found to discover better next time. Every row of Table~\ref{tab:comparison} (Appendix~\ref{app:refarch-audit}), ours included up to this point, backtests only that \emph{output}: discover a factor set once, hold out a test window, report what that one set did. This licenses a claim about the factor set, not the process that produced it: a process can get lucky once on a fixed split, invisibly to the output alone. It is also a different evaluation target than \S\ref{sec:method-standard}'s: there, each admitted factor is scored on its own information coefficient and Sharpe; here, we score the \emph{pool}---the combinator's portfolio over the full admitted library, what a deployed system actually trades, not any one signal alone.

We propose backtesting the \emph{discovery system} itself---supplying the point-in-time process principle (P2, \S\ref{sec:principles}) no system we survey satisfies: re-executing the entire loop of Figure~\ref{fig:arch} at a sequence of historical decision dates, each conditioning only on a sliding in-sample window ending at that date. Concretely, this means running a \textbf{rolling} protocol: starting from an identical library and gate, the system additionally reproposes---new seeds, new proposals, a re-run gate, growing the library---every $k$ months, with the combinator refit on a finer, independent cadence every $j$ months, $j$ dividing $k$ so a newly admitted factor is reflected in the portfolio by the next refit rather than sitting un-weighted until the next repropose. A \textbf{static} backtest---discover once, thereafter only re-fit the combinator on the frozen library as new months arrive---is what every system in Table~\ref{tab:comparison} runs by default, and all a system with no re-executable loop can ever produce: sufficient for backtesting an output, not for backtesting the system that produced it.

Re-executing the loop is necessary but not sufficient: a rolling arm that discarded all state and reproposed cold at every event would just be several independent static runs stitched together. What makes rolling re-discovery a test of the \emph{system}'s adaptivity, not resampling luck, is that SEADS carries state across repropose events---the growing library becomes tomorrow's building blocks (Figure~\ref{fig:arch}), the novelty reference set updates to include newly admitted factors, and lessons persist via the reflector (\S\ref{sec:seads-state}). A system with no carry-forward can still be rolled, but only memorylessly, which tests a strictly weaker claim.

\section{Experimental Setup}
\label{sec:setup}

\paragraph{Two comprehensive US equity panels.} We evaluate on two panels posing different discovery questions. \textbf{Panel A (JKP)} is a large, already heavily-mined characteristics panel---the US subset of JKP's globally-constructed database \citep{jensenkellypedersen2023} (ex-microcap)---checking novelty against that same curated zoo. \textbf{Panel B (CRSP/Compustat base features)} instead gives the system primitive, uncurated accounting and market variables, checking novelty against an external published reference \citep{bryzgalova2025missing}. Both use an in-sample window of 2010--2019; Panel A's out-of-sample window runs to December 2025 ($n\approx71$ months) and Panel B's to November 2024 ($n\approx59$ months), reflecting each source's data availability (\S\ref{sec:limits}). Appendix~\ref{app:data} gives full construction details.

\paragraph{Five re-implemented baselines.} We re-implement AlphaAgent, AlphaForge, RD-Agent(Q), QuantaAlpha, and Beyond Prompting \citep{alphaagent2025,alphaforge2025,rdagentq2025,quantaalpha2026,beyondprompting2026} rather than cite each system's own number on its native market, which would not isolate method from data. Porting fidelity is not uniform, so results below are our reimplementations, not the named systems' own published numbers: AlphaForge and QuantaAlpha are verified against released source; RD-Agent(Q) against the official package; AlphaAgent and Beyond Prompting are close reconstructions from paper text, the latter releasing no code. Every system proposes on the same in-sample window at a matched 300-candidate budget, scored out-of-sample by identical code, using the same current LLM backbone (\S\ref{sec:leakage}); Appendix~\ref{app:baselines} details each port.

\paragraph{Validation gate and backtesting-the-system protocol.} SEADS uses one locked \emph{internal} gate across both panels---coverage, Rank-IC, out-of-decade regime, novelty, partial-IC, and Fama--MacBeth thresholds, in \S\ref{sec:seads-gate}'s order and at the values given in Appendix~\ref{app:metrics}. Each threshold is either a fixed, discipline-conventional constant (e.g.\ novelty's 0.7 cutoff) or governed by a fixed rule as a function of cumulative trial count (the Bonferroni-dynamic Rank-IC bar, Appendix~\ref{app:metrics})---the rule itself set ex-ante and never adjusted after observing a run's OOS performance. Each baseline retains its own native admission logic (Appendix~\ref{app:baselines}); every system's factors are then scored identically by \S\ref{sec:method-standard}'s external standard, not SEADS's gate. For \S\ref{sec:method-backtest}'s protocol, we evaluate an annual re-discovery cadence with a 4-month combinator refit, against an identical-library static baseline; this cadence and \S\ref{sec:ablations}'s ablation conditions were likewise fixed pre-run from in-sample evidence alone.

\section{Results}
\label{sec:results}

\subsection{Main Out-of-Sample Comparison}
\label{sec:main}
\begin{table}[t]
\centering
\small
\tabcolsep=2pt
\renewcommand{\arraystretch}{0.94}
\begin{tabular}{@{}L{2.9cm}cccc@{}}
\toprule
System & Productivity & Performance & Novelty & Spanning \\
 & (A\,/\,B) & (A\,/\,B) & (A\,/\,B) & (A\,/\,B) \\
\midrule
AlphaForge & 5.0/1.8 & 0.34/$-0.08$ & 0.96/0.74 & $-0.13$/0.56 \\
RD-Agent(Q) & 39.0/47.0 & 0.19/$-0.01$ & 0.79/0.64 & 1.27/0.88 \\
AlphaAgent & 31.0/32.6 & 0.12/0.09 & 0.82/0.67 & $-0.76$/0.34 \\
QuantaAlpha & 60.8/34.6 & $-0.01$/0.04 & \textbf{0.65}/0.61 & 0.37/\textbf{1.46} \\
Beyond Prompting & \textbf{96.4}/\textbf{85.0} & \textbf{0.41}/0.11 & 0.90/0.81 & \textbf{2.04}/1.01 \\
\textbf{SEADS (ours)} & 14.0/13.8 & 0.25/\textbf{0.16} & \textbf{0.65}/\textbf{0.46} & 1.21/1.28 \\
\bottomrule
\end{tabular}
\caption{Matched-budget, out-of-sample-only comparison (P3, \S\ref{sec:principles}); identical scoring on both panels (\S\ref{sec:setup}); A\,=\,Panel A (JKP), B\,=\,Panel B (CRSP/Compustat); mean across 5 replicates; per-entry cross-replicate SEs in Table~\ref{tab:main-se} (Appendix~\ref{app:metrics}). \textbf{Productivity}, \textbf{Performance}, \textbf{Novelty}, and \textbf{Spanning Test} are defined in \S\ref{sec:method-standard}; higher is better for all but Novelty, where lower means more novel. \textbf{Bold} marks each column's best value per panel.}
\label{tab:main}
\end{table}

Table~\ref{tab:main} reports productivity, performance, novelty, and a spanning test for all six systems, each entry a mean over 5 replicates. Three rankings disagree, and the winner shifts by panel: by performance, Beyond Prompting leads Panel A (0.41, AlphaForge second at 0.34, SEADS third at 0.25) while SEADS leads Panel B (0.16 vs.\ 0.11, AlphaForge last at $-0.08$); by novelty, SEADS and QuantaAlpha are tied atop Panel A (0.65 each, AlphaForge worst of all six at 0.96), while SEADS leads Panel B outright (0.46 vs.\ QuantaAlpha's next-best 0.61); by productivity, Beyond Prompting leads both panels (96.4/85.0 vs.\ SEADS's 14.0/13.8). AlphaForge's own split---strong performance, least novel of any system---repeats \S\ref{sec:method-standard}'s productivity argument on the novelty axis: performance is cheap without novelty. Only on Panel B does one system, SEADS, rank top two on both novelty and performance; on Panel A none does, since the two performance leaders (Beyond Prompting, AlphaForge) are also the two least novel. Beyond Prompting's productivity and performance both partly reflect a looser gate and predominantly redundant formulas (Appendix~\ref{app:complexity-case}), not just superior discovery. No single static ranking is decisive: \S\ref{sec:method-backtest} asks a different question of the same systems, not a fourth static one.

Panel B's looser reference set makes every system more novel (Table~\ref{tab:main}), while its weaker raw features lower mean performance---least for SEADS, whose admits are mostly multiplicative interactions rather than the additive spreads RD-Agent(Q) and Beyond Prompting favor (Appendix~\ref{app:complexity-case}).

\subsection{A Rolling Comparison}
\label{sec:wf}
\begin{figure}[t]
\centering
\includegraphics[width=0.88\columnwidth]{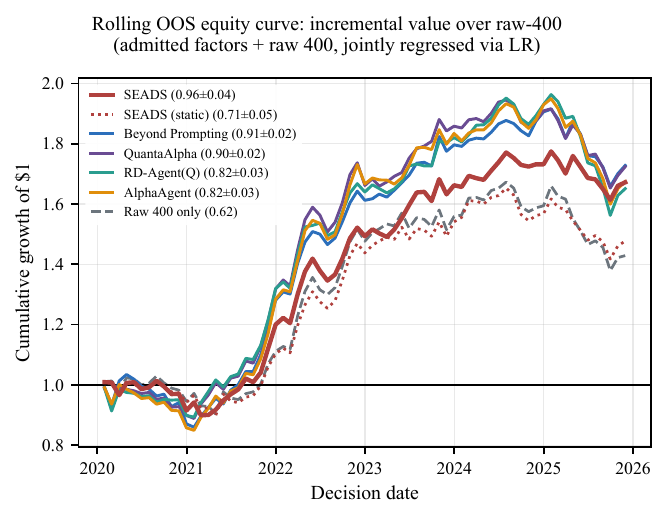}
\caption{\S\ref{sec:method-backtest}'s protocol applied on Panel A (JKP): cumulative OOS growth of \$1 under SEADS and the four other agentic factor-discovery baselines. Each system's admitted factors are jointly regressed with the full raw $\sim$400-characteristic panel, not combined alone, so an admitted-only Sharpe edge from a large, redundant set (Appendix~\ref{app:complexity-case}) cannot be mistaken for value the raw panel already supplies; \textbf{Raw 400 only} is the resulting no-discovery floor: the same raw panel, refit identically, with no admitted factors, fully deterministic (no LLM step, no replicate variance). \textbf{SEADS (static)} isolates repropose's contribution: its own starting library and refit schedule, never re-discovering. The line shown for each of the other six is one of 3 replicates; the legend's Sharpe is instead the mean $\pm$ SE across all 3, each computed independently before averaging, so a line's equity path need not track its own legend ranking---Sharpe rewards steadier growth, not a higher endpoint. AlphaForge is omitted (too few admits).}
\label{fig:wf-main}
\end{figure}
Figure~\ref{fig:wf-main}'s clearest pattern is that all five discovery systems clear the no-discovery floor by a wide margin (mean Sharpe 0.82--0.96 vs.\ Raw 400 only's 0.62): whatever each system's admitted factors contribute on top of the raw panel, it substantially exceeds what a plain linear model already extracts from that panel alone. Among the five, SEADS nominally leads (0.96$\pm$0.04) but is not reliably distinguishable from Beyond Prompting (0.91$\pm$0.02) or QuantaAlpha (0.90$\pm$0.02) given cross-replicate noise; it is more clearly separated from RD-Agent(Q) and AlphaAgent (0.82$\pm$0.03 each). This compares each system's best rolling configuration, not an isolated persistence effect---SEADS's carry-forward arm and each baseline's own rolling arm still differ in algorithm, gate, and reconstruction fidelity, not persistence alone. SEADS's static arm---identical starting library, never re-discovering---reaches only 0.71$\pm$0.05, closer to the no-discovery floor than to any rolling system: most of SEADS's rolling advantage is repropose itself, not the initial library.

\subsection{Ablations}
\label{sec:ablations}
\begin{table}[t]
\centering
\small
\tabcolsep=3pt
\begin{tabular}{@{}L{3.2cm}ccc@{}}
\toprule
Configuration & Productivity & Performance & Novelty \\
\midrule
Full pipeline & $14.0\pm1.4$ & $0.254\pm0.03$ & $0.653\pm0.01$ \\
Planner off & $37.2\pm2.5$ & $0.101\pm0.04$ & $0.676\pm0.01$ \\
Align Check off & $17.6\pm1.2$ & $0.215\pm0.03$ & $0.662\pm0.01$ \\
Mem (Lessons) off & $17.2\pm1.0$ & $0.203\pm0.03$ & $0.650\pm0.01$ \\
Mem (Attempts) off & $10.0\pm0.6$ & $0.096\pm0.02$ & $0.627\pm0.01$ \\
Self-evolution off & $7.8\pm0.6$ & $0.080\pm0.02$ & $0.630\pm0.02$ \\
\bottomrule
\end{tabular}
\caption{Panel A mechanism ablations, testing whether each of SEADS's mechanisms is load-bearing: one mechanism removed at a time relative to the full pipeline, mean $\pm$ SE across 5 replicates. ``Memory (lessons/attempts) off'' isolates the reflector's natural-language lessons from the append-only attempt log (\S\ref{sec:seads-state}). Higher is better for productivity and performance; lower is more novel.}
\label{tab:ablation-main}
\end{table}
We run a battery of ablations (Table~\ref{tab:ablation-main}), each isolating one mechanism relative to the full pipeline. Self-evolution is the single most load-bearing mechanism: removing it shrinks the admitted set itself (7.8 vs.\ 14.0 admits) and cuts Sharpe by 68\% (0.08 vs.\ 0.25), the largest drop in the battery. Removing the Planner instead inflates admissions nearly threefold (37.2 vs.\ 14.0) while Sharpe still collapses (0.10)---a different failure mode, too many weak factors rather than too few. Removing the Alignment Checker or either memory tier alone costs 15--62\% of Sharpe, while novelty moves comparatively little across the whole battery. The validation gate and look-ahead firewall are audited differently: turning either off changes what counts as an admission at all, not just how many, so neither is ablated outright.

\section{Limitations}
\label{sec:limits}
\emph{Small samples}: admitted-factor counts run in the tens per replicate, OOS windows 59--71 months, so most cross-system differences are not reliably distinguishable. Replicate counts (5 static, 3 rolling) reflect this study's API budget, not a deliberate power limit.

\emph{Independent admission does not guarantee a coherent combined portfolio}: the gate certifies a factor's marginal in-sample signal, not its joint, non-monotone out-of-sample contribution alongside later admits (Figure~\ref{fig:count}, Appendix~\ref{app:internals}).

\emph{Achievable Sharpe is dataset-dependent}: the panels differ in alpha headroom, so absolute Sharpe alone is not informative---\S\ref{sec:main} compares systems within a panel instead.

\label{sec:leakage}\emph{Backbone information leakage is soft, not absent}: SEADS uses one current LLM (GPT-5.2) for every role, not a release-dated model per historical date. Proposing from feature \emph{descriptions} rather than realized returns keeps this close to the safe end, but a soft residual remains: general knowledge of which factor families worked could still bias proposed hypotheses \citep{profitmirage2025,glassermanlin2023,sarkarvafa2025,memorization2025}, more so across the rolling protocol's several decision dates than the static comparison's single cutoff. A release-dated backbone (e.g.\ GPT-4o-mini, October 2023) is the natural alternative, but leaves only $\sim$25 clean OOS months---too few here, and confounded with a weaker, older model's reasoning capacity.

\emph{Our look-ahead firewall may be overly conservative}: static pattern-matching can reject safe formulas with unsafe ones; a less conservative alternative exists but postdates these results.

\emph{Candidate counts are matched, compute is not}: the 300-candidate budget matches initial proposals, but self-evolution's mutation retries and repairs add LLM calls beyond one proposal per slot---removing it alone collapses Sharpe by 68\% (Table~\ref{tab:ablation-main})---so realized compute per admission is not matched the way candidate counts are.

\emph{Reported Sharpe is gross of trading costs}: we model neither transaction costs nor turnover, so frequent reranking or more complex formulas could erode realized performance disproportionately, a gap no number reported here reflects.

\section{Conclusion}
\label{sec:conclusion}
AEAP has neither a shared definition nor a shared evaluation discipline for systems that discover their own hypotheses rather than test a researcher's. Our three contributions (\S\ref{sec:intro}) give a definition and its building blocks (C1); an output-facing standard scoring a discovered set on multiple axes jointly, plus a method for backtesting the discovery \emph{system} across historical decision dates rather than just one run's output (C2); and SEADS, instantiating both against five re-implemented baselines (C3).

Across both panels, no system excels on every standard \S\ref{sec:method-standard} scores; even a multi-axis standard cannot settle the best system from one decision date alone. \S\ref{sec:method-backtest} asks what static scoring cannot: whether the discovery process holds up when re-executed under historical, point-in-time constraints. Our two negative findings---non-monotone combination, a soft backbone-leakage residual---are evidence an unsettled field should expect, not a failure of one system.

\section*{Ethical Statement}
This work concerns automated discovery of trading signals, a dual-use capability whose primary risk is overstated performance claims rather than any single dangerous capability. Our evaluation standard is deliberately deflationary: scoring a discovery system out-of-sample and reporting negative findings reduces, rather than inflates, the field's headline claims. Automated discovery at scale can contribute to crowding if deployed identically by many actors; our panels include small-cap names with capacity limits our construction does not model. Nothing here is investment advice.

\clearpage
\small
\bibliography{references}

\clearpage
\appendix
\begin{center}\Large\textbf{Appendix}\end{center}
\noindent This appendix gives full methodology and results supporting the main paper, which cites specific tables/figures below where relevant.

\section{Data Construction Detail}
\label{app:data}

\paragraph{Panel A: JKP characteristics (ex-microcap).} Built from the JKP global factor characteristics data \citep{jensenkellypedersen2023}, restricted to US-listed stocks (\texttt{excntry = USA}). Four size-filtered variants of this panel exist in our data pipeline; the one used for every result in this paper drops rows whose \texttt{size\_grp} label (recomputed every month from that month's NYSE market-cap percentile breakpoints, so the cut itself carries no look-ahead) is \texttt{micro} or \texttt{nano}, retaining \texttt{small}, \texttt{large}, and \texttt{mega}. This yields 1{,}936{,}944 stock-months across 21{,}052 unique stocks, 1925--2025, and $\sim$400 taxonomy-declared characteristics (representative examples: one-year asset growth, book-to-market, operating-cash-flow-to-assets, gross profitability, sales growth, capex intensity, the Piotroski F-score, the Ohlson O-score, the Altman Z-score, 60-month beta, 252-day idiosyncratic volatility, 126-day turnover, and 12-1 momentum). Novelty on Panel A is checked against this same panel, so a candidate must be distinct from the already-curated JKP zoo itself, not merely from a smaller reference. The in-sample window used for every result in this paper is 2010--2019; the out-of-sample window runs through December 2025 ($n\approx71$ months).

\paragraph{Panel B: CRSP/Compustat base features.} Two further panels support Panel B. The \emph{base-features} panel is built directly from CRSP and Compustat as 87 comparatively primitive accounting and market variables, organized into 13 categories (trading frictions, market price, size, operating assets, liquid assets, intangibles, leverage, equity structure, revenue and costs, cash flow, earnings, dividends, and capital actions; category sizes range from 3 to 13 variables): 1{,}213{,}574 stock-months, 5{,}907 unique stocks, 1975--2024. These are deliberately \emph{not} pre-curated characteristics---ratios and transforms an existing factor library would already apply are left for the discovery system itself to construct.

\paragraph{Panel B's novelty reference.} The \emph{reference} panel for novelty on Panel B is the 45-characteristic panel of \citet{bryzgalova2025missing} (``Missing Financial Data''), built by that paper's own local cross-sectional/backward-cross-sectional imputation method without look-ahead, restricted to the top 80\% of stocks by the previous month's market equity: 2{,}402{,}623 stock-months, 22{,}336 unique stocks, 1977--2024, spanning six categories (value, profitability, investment, trading frictions, past returns, and a single size/intangibles variable; category sizes 1--13). We check Panel B's discovered factors against this established reference rather than against Panel B's own 87 raw inputs, since the latter would reward literal repackaging of a single input as ``novel.'' The in-sample window matches Panel A (2010--2019); the out-of-sample window runs through this panel's own data availability, November 2024 ($n\approx59$ months)---shorter than Panel A's, correspondingly lowering this panel's statistical power (\S\ref{sec:limits}).

\paragraph{Point-in-time convention.} Every feature source in our data layer is contractually required to store its return column as a lead-1-month return---$\mathrm{ret}(t)$ realized over $(t,t{+}1]$---documented in each source's taxonomy entry rather than left to each track's own convention. This is a real invariant, not a formality: a per-track lag parameter was tried and removed in favor of the uniform contract, since a single wrong return-column convention on any one panel silently corrupts every factor built on top of it.

\section{Reference Architecture Component Audit}
\label{app:refarch-audit}
Table~\ref{tab:comparison} summarizes what each of the five re-implemented systems' own paper reports about itself (\S\ref{sec:setup} re-implements and re-scores all six under one shared protocol instead). Table~\ref{tab:refarch-audit} then audits which components of Figure~\ref{fig:arch} each system instantiates, expanding Table~\ref{tab:comparison}'s summary columns into four of \S\ref{sec:aap-def}'s five building blocks (B3, execution against point-in-time data, is omitted as uniform across every factor-discovery system surveyed) plus backtesting-the-system.

\begin{table}[t]
\centering
\small
\tabcolsep=3pt
\begin{tabular}{@{}L{2.9cm}L{2.9cm}cc@{}}
\toprule
System & Validation signal & Nov.\ check? & \# gates \\
\midrule
AlphaAgent & AST-originality + LLM judge & \checkmark & 2 \\
AlphaForge & fitness pools & \ding{55} & 1 \\
RD-Agent(Q) & backtest-metric unit & \ding{55} & 1 \\
QuantaAlpha & LLM-judged semantic + complexity & partial & 2 \\
Beyond Prompting & pre-spec.\ stat.\ gates & \ding{55} & 2 \\
\midrule
\textbf{SEADS (ours)} & \textbf{multi-stage stat.\ gate} & \textbf{\checkmark} & \textbf{9} \\
\bottomrule
\end{tabular}
\caption{Per-system comparison, based on what each cited system's own paper reports about itself. Whether the discovery loop is itself re-executed at more than one historical decision date (\S\ref{sec:method-backtest}) is audited separately in Table~\ref{tab:refarch-audit}. ``\# gates'': count of independent admission criteria, statistical or heuristic.}
\label{tab:comparison}
\end{table}
\begin{table}[t]
\centering
\footnotesize
\tabcolsep=2pt
\begin{tabular}{@{}L{2.6cm}L{2.2cm}L{2.6cm}L{1.8cm}L{2.8cm}c@{}}
\toprule
System & B1 Hyp.\ & B2 Formal.\ & B4 Gate & B5 Persist.\ & Backtests \\
\midrule
AlphaAgent & LLM idea agent & symbolic factor & 2 criteria & AST-orig.\ memory & \ding{55} \\
AlphaForge & none (mined) & symbolic factor & fitness pool & factor zoo & \ding{55} \\
RD-Agent(Q) & LLM hypothesis & free-form code & 1 unit & knowledge forest & \ding{55} \\
QuantaAlpha & div.\ planning & symbolic factor & 2 criteria & trajectory pop.\ & \ding{55} \\
Beyond Prompting & closed-loop agent & symbolic factor & 2 (pre-spec.) & strategy update & \ding{55} \\
\textbf{SEADS} & \textbf{Planner + Proposer} & \textbf{Python/pandas} & \textbf{9 criteria} & \textbf{tax.+log+lessons+lib.} & \textbf{\checkmark} \\
\bottomrule
\end{tabular}
\caption{Reference-architecture component audit (\S\ref{sec:refarch}), based on our re-implementation of each system (\S\ref{sec:setup}, Appendix~\ref{app:baselines}). AlphaForge's ``none (mined)'' B1 cell reflects that it is not itself an AEAP system by \S\ref{sec:aap-def}'s definition (no LLM-driven hypothesis step)---we include it as the field's standard non-agentic comparison baseline, not as evidence that the building blocks are optional within the AEAP systems we survey.}
\label{tab:refarch-audit}
\end{table}

\section{The SEADS Discovery Loop: Pseudocode}
\label{app:algorithm}
Algorithm~\ref{alg:seads} states \S\ref{sec:seads}'s five named components (Figure~\ref{fig:arch}) as one closed loop, at the level of abstraction Table~\ref{tab:refarch-audit} compares systems at. It elides engineering detail---concurrency, name-collision handling, exact prompt content---covered instead in Appendix~\ref{app:internals}; \textsc{Gate} is Table~\ref{tab:gate}'s full ordered sequence, not a single test, and \textsc{SelfEvolve} re-enters at Step 2 for the same slot rather than starting a new one. The rolling protocol (Appendix~\ref{app:wf-protocol}) re-invokes this whole procedure at each repropose event as walk-forward time advances, rather than nesting that outer loop here.

\begin{algo}[t]
\small
\caption{SEADS Closed-Loop Factor Discovery}
\label{alg:seads}
\begin{algorithmic}[1]
\Require Point-in-time panel $\mathbf{X}$; reference set $\mathcal{Z}$; gate thresholds $\Theta=\{\tau_{cov},\tau_{ic},\tau_{regime},\tau_{nov},\tau_{pic},\tau_{fm}\}$; rounds $K$; slots per round $S$, $K{\times}S{=}300$; max self-evolution attempts $E$
\Ensure Admitted factor library $\mathcal{G}$; combinator weights $w$
\State $\mathcal{G} \gets \emptyset$; memory $(T_1,T_2,T_3) \gets (\emptyset,\emptyset,\emptyset)$
\For{$round = 1$ \textbf{to} $K$}
  \State $\tau_{ic} \gets$ Bonferroni-dynamic update from trials so far \Comment{Appendix~\ref{app:metrics}}
  \Statex \textbf{\quad Step 1: Generation (Planner, Proposer)} \Comment{\S\ref{sec:seads-gen}}
  \State $\{seed_1,\ldots,seed_S\} \gets$ \textsc{Planner}$(T_1,T_2,T_3,\mathcal{G})$ \Comment{one call plans the round}
  \For{$slot = 1$ \textbf{to} $S$ \textbf{(independent; dispatched concurrently)}}
    \State $\theta \gets$ \textsc{Proposer}$(seed_{slot}, T_1,T_2,T_3,\mathcal{G})$ \Comment{rationale + symbolic formula}
    \For{$attempt = 0$ \textbf{to} $E$}
      \If{$attempt > 0$}
        \State $\theta \gets$ \textsc{SelfEvolve}$(\theta, stage, T_2)$ \Comment{\textbf{Step 4} (\S\ref{sec:seads-evolve}): mutate the near-pass}
      \EndIf
      \Statex \textbf{\quad\quad Step 2: Sanity Checks} \Comment{\S\ref{sec:seads-sanity}}
      \If{\textbf{not} \textsc{StructurallyValid}$(\theta)$}
        \State \textbf{break} \Comment{malformed spec; no code executes}
      \EndIf
      \State $code \gets$ \textsc{Implement}$(\theta, \mathbf{X})$ \Comment{sandboxed; firewall + $\le2$ LLM repairs}
      \If{firewall fails \textbf{or not} \textsc{AlignmentCheck}$(\theta, code)$}
        \State \textbf{break}
      \EndIf
      \Statex \textbf{\quad\quad Step 3: Validation Gate} \Comment{\S\ref{sec:seads-gate}}
      \State $decision, stage \gets$ \textsc{Gate}$(f_\theta, \mathbf{X}, \mathcal{G}, \mathcal{Z}, \Theta)$ \Comment{sequential AND, Table~\ref{tab:gate}}
      \If{$decision = $ \emph{admit}}
        \State $\mathcal{G} \gets \mathcal{G} \cup \{f_\theta\}$; \textbf{break} \Comment{\S\ref{sec:seads-state}}
      \ElsIf{\textbf{not} \textsc{NearPass}$(stage)$ \textbf{or} $attempt = E$}
        \State \textbf{break} \Comment{far short of any bar, or evolution budget spent}
      \EndIf
    \EndFor
  \EndFor
  \Statex \textbf{\quad Step 5: Persistent State} \Comment{\S\ref{sec:seads-state}}
  \State update $T_1$ (characteristic stats), $T_2$ (every attempt and its $stage$); $T_3 \gets$ \textsc{Reflector}$(T_2)$
\EndFor
\State $w \gets$ \textsc{Combinator}$(\mathcal{G})$ \Comment{Ridge, $\alpha{=}1$, rank-normalized library values}
\State \Return $\mathcal{G}, w$
\end{algorithmic}
\end{algo}
\textsc{NearPass} holds when $\theta$ cleared a minimum progress floor or failed specifically at the rank-IC or partial-IC stage (Appendix~\ref{app:internals}); a candidate failing far short of any bar is discarded rather than mutated. Every system in Table~\ref{tab:comparison} instantiates Steps 1--3 in some form, and Step 5's cross-round persistence appears in every system too (Table~\ref{tab:refarch-audit}'s B5 column), though SEADS's own mechanism is the most elaborate (four components vs.\ one each); Step 4---mutating a near-miss rather than discarding it---is not something Table~\ref{tab:refarch-audit} audits for any system.

\section{Validation Gate and Firewall Detail}
\label{app:gatefirewall}
Table~\ref{tab:firewall} gives the look-ahead firewall's rule classes; Table~\ref{tab:gate} gives the validation gate's exact current execution order, expanding \S\ref{sec:seads-gate}. The gate is deterministic throughout; admission is the logical AND of every configured check, short-circuiting at the first failure, followed by the sign backstop.

\begin{table}[t]
\centering
\small
\tabcolsep=3.5pt
\begin{tabular}{@{}cL{1.8cm}L{4.6cm}@{}}
\toprule
\# & Rule class & What it blocks (static AST) \\
\midrule
1 & Forbidden columns & any read of return/target columns via subscript, attribute, or \texttt{.get} \\
2 & Dynamic access & \texttt{getattr}, \texttt{eval}, \texttt{exec}, \texttt{.eval()}, \texttt{.query()} \\
3 & Negative shift & \texttt{.shift/.diff/.pct\_change} with a negative literal \\
4 & Centered window & \texttt{.rolling(center=True)} \\
5 & Bare rank & \texttt{.rank()} not scoped to \texttt{groupby('date')} \\
6 & Per-stock transform & \texttt{groupby('permno')}\newline\texttt{.transform(...)} over full history \\
7 & Full-panel/expand & panel-wide \texttt{mean/std/\ldots} or \texttt{expanding/cumsum} without a safe receiver \\
\bottomrule
\end{tabular}
\caption{The seven rule classes of the deterministic AST look-ahead firewall; a whitelist additionally enforces that every column read is taxonomy-declared, and every read is further scoped to a sandboxed namespace exposing only the formula's statically-read columns.}
\label{tab:firewall}
\end{table}

\begin{table}[t]
\centering
\small
\tabcolsep=5pt
\begin{tabular}{@{}lL{2.0cm}L{4.05cm}cL{1.5cm}@{}}
\toprule
\# & Check & Threshold key & Value & Runs if \\
\midrule
0a & duplicate fingerprint & --- & --- & always \\
0b & self-collapse & fixed & 0.95 & always \\
1 & coverage & \texttt{Min\_Coverage} & 0.5 & always \\
2 & Rank-IC $t$-stat & \texttt{Min\_RANKIC\_T} & $\max(3.0,$ Bonf.-dyn.$)$ & coverage ok \\
3 & regime consist.\ & \texttt{Regime\_Lookback\_Years} & 10 & rank-IC ok \\
3.5/4 & novelty & \texttt{Max\_Corr} & 0.7 & 2, 3 ok \\
5 & partial-IC & \texttt{Min\_Partial\_IC} & 1.0 (top-5) & 2, 3, 4 ok \\
6 & Fama--MacBeth & \texttt{Min\_FM} & 1.0 (top-5) & 2, 3, 4, 5 ok \\
7 & sign alignment & determ.\ & --- & alpha-pass \\
\bottomrule
\end{tabular}
\caption{Validation-gate execution order and threshold values (\S\ref{sec:seads-gate}), used for every result in this paper. The self-collapse check (0b) rejects a candidate whose final formula correlates $\ge0.95$ with a plain rank of one of its own declared input variables. The regime-consistency check's (3) window re-anchors to the in-sample start, sliding automatically under \S\ref{sec:method-backtest}'s rolling protocol. The sign check (7) requires the realized rank-IC, partial-IC, and Fama--MacBeth beta to agree with each other in sign. Novelty runs after rank-IC/regime rather than before because the shared control-panel correlation matrix used by checks 4--6 is computed once, unconditionally, the moment rank-IC passes, making novelty a cheap lookup by the time it runs, while it would otherwise be the single slowest check once the reference panel is large (measured at over 100 seconds per candidate against a $\sim$400-column reference, vectorized). The candidate budget (300) is a separate, global experimental parameter, not a per-check threshold.}
\label{tab:gate}
\end{table}

\section{Pipeline Internals}
\label{app:internals}
\paragraph{Concurrency and naming.} Within one run, shared mutable state (the control panel, the trial counter, the name registry) is mutated strictly sequentially or under a lock, so concurrent candidate dispatch cannot corrupt an admission decision; a name collision bumps an existing version suffix rather than stacking a new one.
\paragraph{Self-evolution.} A candidate is mutated only if its rank-IC $t$-statistic cleared a minimum progress floor of 2.0 or it failed specifically at the rank-IC or partial-IC stage; one failing far short of any bar is discarded rather than mutated. A progress-weighted score ranks a mutation strictly by the latest gate stage it reached (rank-IC $<$ regime-consistency $<$ novelty $<$ partial-IC $<$ Fama--MacBeth, each a full tier above the last), with magnitude breaking ties only within a tier, so that reaching further into the gate always outranks a larger margin earlier in it. Its mutations are correlated draws from the same parent candidate rather than fresh independent trials; whether the rank-IC gate's live Bonferroni correction (Appendix~\ref{app:metrics}) should count each mutation separately toward the cumulative candidate count or discount correlated attempts is left for future work.
\paragraph{Planner and Proposer context scope.} The Planner sees the full history of past attempts when planning a new round; each seed's Proposer instance instead sees only prior attempts sharing at least one of that seed's underlying variables, not the Planner's full history. This scoping matches each role's job: the Planner needs breadth across themes and variable families, not formula-construction guidance, since it emits a seed rather than a formula; the Proposer needs depth on a narrow, variable-specific slice of prior attempts, and correspondingly more explicit formula-construction guidance than the Planner requires.
\paragraph{Planner backstops and structural pre-check.} Two deterministic backstops correct observed Planner failure modes: one forces complexity-tier rebalancing when the Planner's own ``aim for a mix'' instruction alone leaves it over-concentrated on a single tier (simple, moderate, or rich); the other substitutes fresh variables into an explore seed whose variable pair was already tried repeatedly, since the Planner otherwise kept re-selecting the same pair despite full visibility into past attempts. A deterministic structural pre-check separately rejects malformed specs before any code executes: naming convention, minimum rationale length, a non-empty formula, and a valid category.
\paragraph{Prompt design.} Of SEADS's six LLM-prompted roles (Table~\ref{tab:llmconfig}), four---Planner, Proposer, Self-Evolve, Alignment Checker---share one structural convention: a \texttt{GENERAL INFO} block states the role's job and previews the prompt's own section order (e.g.\ the Planner's own preamble: ``RULES you must follow, then three memory tiers\ldots then FINAL DIRECTION with the exact output format''), so a long, many-section prompt is self-documenting before the model reads past the first paragraph. \texttt{RULES} follow next; the Proposer and Self-Evolve additionally interleave \texttt{PANEL STRUCTURE \& LOOK-AHEAD RULES} and \texttt{FACTOR QUALITY GUIDANCE} here, since panel conventions and what makes a formula worth proposing are each several distinct points, not one rule among others---the Planner has neither, since it emits a seed rather than a formula. The three memory tiers---Building Blocks ($T_1$), Past Attempts ($T_2$), Lessons Learned ($T_3$) in Algorithm~\ref{alg:seads}'s notation---come after; the Alignment Checker has none, since it judges one already-written formula in isolation rather than proposing a new one. A mandatory pre-submission self-check and the output template both sit last, immediately before generation, on the same closest-instruction-wins logic. Implementation Agent and the reflector use simpler, single-shot prompts scoped to their narrower tasks (fixing a formula; summarizing a round into a lesson), not this multi-section structure.
\paragraph{Alignment Checker grounding.} The Alignment Checker's prompt requires a \texttt{formula\_trace}---a mechanical, line-by-line restatement of what each step computes, filled in before the aligned/not-aligned verdict---for two purposes. First, grounding: the checker once rejected a formula by citing a \texttt{.shift(1)} call that did not appear anywhere in the actual code, a fabrication that cost a pipeline slot on a formula with no real defect, so every look-ahead rejection now must quote an exact substring present in that item's own formula. Second, algebraic cancellation, motivated by a second real incident: a formula computing \texttt{(shrout*cfacshr)/shrout} was live and admitted before this check existed, algebraically just \texttt{cfacshr}---\texttt{shrout} cancels exactly---even though its rationale described a genuine two-variable interaction between share count and an adjustment factor; the trace must now catch a formula that reduces to fewer effective variables than \texttt{variables\_used} claims, since that misdescribes what the formula actually computes regardless of how genuine the stated interaction sounds. Rank direction and sign convention are explicitly out of scope for this judge for a separate, sign-specific reason: it had repeatedly rejected correctly-signed formulas for ``not inverting the rank'' when the rationale had already established the raw variable as higher-is-better, so sign checking was moved out of the LLM judge entirely and made a deterministic post-hoc comparison against the realized Fama--MacBeth coefficient instead (\S\ref{sec:seads-gate}, check 7).

\begin{figure}[h]
\centering
\includegraphics[width=0.85\columnwidth]{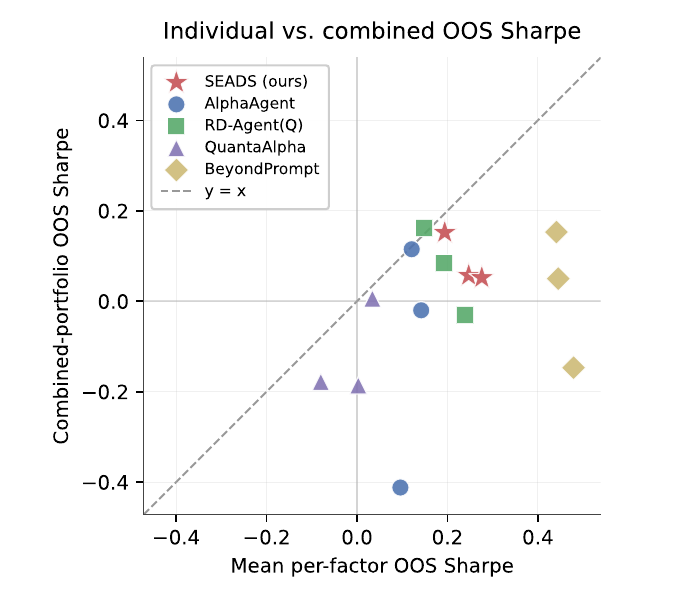}
\caption{Mean per-factor out-of-sample Sharpe vs.\ combined-portfolio out-of-sample Sharpe, three replicates per system, Panel A (JKP). Combined SR is from a single OLS regression of next-month return on the replicate's admitted factors, fit once over the in-sample window and applied out-of-sample to score and rank stocks into a long-short portfolio each month; we report that portfolio's annualized Sharpe. 14 of 15 replicates fall below the dashed diagonal---combining admitted factors typically does not preserve, let alone improve on, their average individual quality (\S\ref{sec:limits}).}
\label{fig:count}
\end{figure}

\begin{figure}[h]
\centering
\includegraphics[width=0.7\textwidth]{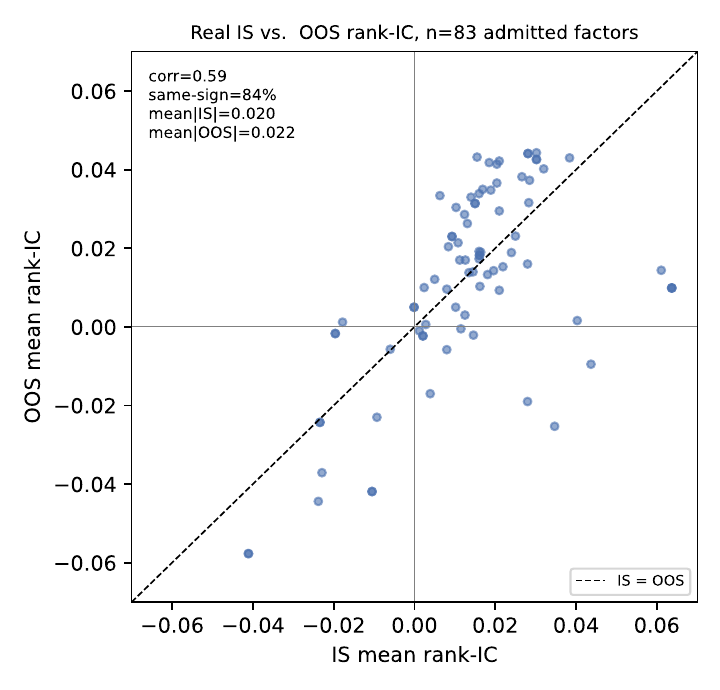}
\caption{In-sample vs.\ out-of-sample mean rank-IC across admitted factors from a few sample runs of SEADS: sign-agreement most of the time, no systematic shrinkage in mean $|$IC$|$, but far from a tight relationship (\S\ref{sec:limits}).}
\label{fig:shrink}
\end{figure}

\paragraph{Combinators.} We use a Ridge regression ($\alpha=1$) fit on the admitted library's rank-normalized values alone, an ex-ante default chosen independently of any comparative performance check. A more expressive learner---XGBoost or LightGBM, say---could also convert the admitted library into scores; we use linear/Ridge instead because it keeps the combinator stable, comparable across systems and panels, and exactly reproducible, properties this evaluation needs more than whatever incremental predictive power a more flexible fit might add. This is not merely a stability preference: every admitted factor already cleared the gate's Rank-IC and partial-IC checks (\S\ref{sec:seads-gate}), both rank/linear-correlation measures, so the library a combinator scores is pre-selected for linear predictive power---a nonlinear combinator has comparatively little nonlinearity left to exploit, and would mainly add tuning variance the evaluation does not need. This deployed Ridge combinator, Figure~\ref{fig:count}'s single-OLS illustration, and the rolling comparison's raw-panel joint regression (Figure~\ref{fig:wf-main}) are three distinct procedures answering three different questions about the same admitted library. Figure~\ref{fig:count} and Figure~\ref{fig:shrink} give two further empirical patterns behind \S\ref{sec:limits}'s claims, and Table~\ref{tab:map} maps every mechanism named above to the module that implements it.
\begin{table}[t]
\centering
\small
\tabcolsep=4pt
\begin{tabular}{@{}L{2.7cm}L{4.9cm}@{}}
\toprule
Mechanism & Module / function \\
\midrule
Planner (seeds) & \texttt{planner\_agent.plan\_run} \\
Proposer & \texttt{hypothesis\_agent.}\newline\texttt{propose\_factor} \\
Look-ahead firewall & \texttt{implementation\_agent.}\newline\texttt{check\_formula\_firewall} \\
Sandboxed execution & \texttt{implementation\_agent.}\newline\texttt{compute\_factor} \\
Alignment / sign & \texttt{rationale\_gate} \\
Validation gate & \texttt{validation\_engine.}\newline\texttt{validate\_factor} \\
Self-evolution & \texttt{self\_evolve\_agent} \\
Memory & \texttt{memory\_manager} \\
Rolling/backtesting-the-system & \texttt{rolling\_engine} \\
Combination & \texttt{combinator} \\
\bottomrule
\end{tabular}
\caption{Map from named mechanisms to the modules that realize them.}
\label{tab:map}
\end{table}

\section{Backtesting-the-System: Full Walk-Forward Protocol}
\label{app:wf-protocol}
\begin{figure}[h]
\centering
\includegraphics[width=\columnwidth]{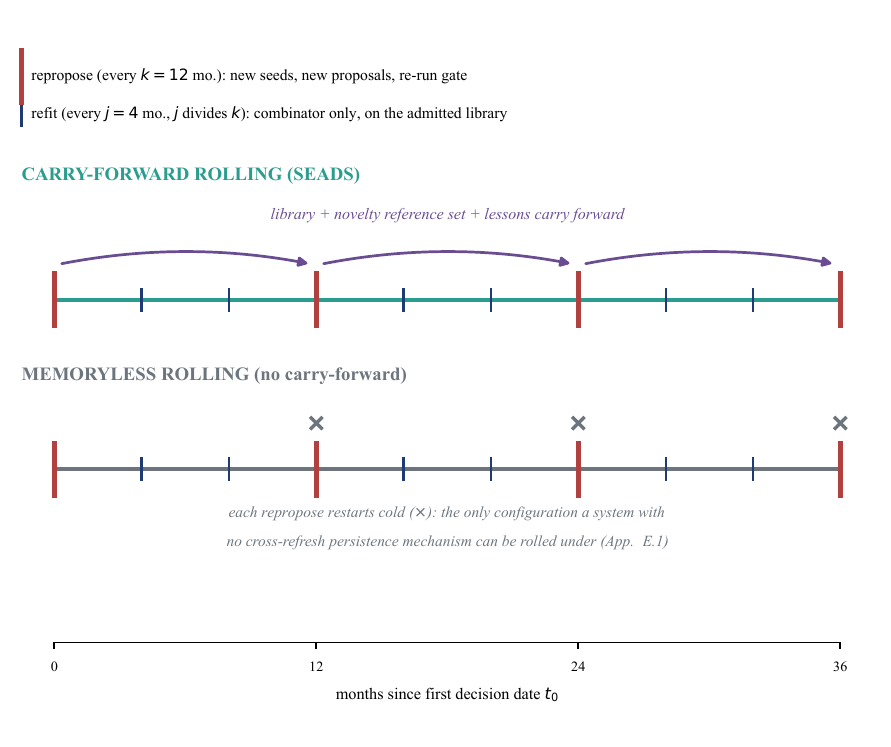}
\caption{Illustrative repropose/refit timeline ($k=12$, $j=4$; \S\ref{sec:method-backtest}), not the actual OOS schedule of \S\ref{sec:setup}. Top: SEADS's carry-forward rolling---library, novelty reference set, and lessons persist across every repropose event. Bottom: baseline rolling---each repropose event re-invokes the baseline's own driver at the new decision date, without SEADS's specific cross-refresh mechanisms (library, novelty reference set, reflector lessons). Both rows share the identical repropose/refit tick structure; only what crosses between ticks differs.}
\label{fig:wf-timeline}
\end{figure}
We use ``rolling'' and ``walk-forward'' interchangeably for \S\ref{sec:method-backtest}'s protocol, since it is the same sliding-window re-evaluation discipline standard in trading-strategy validation, applied here to the discovery process itself rather than to a fixed model's parameters. The engine supports two factor-update modes at each re-discovery event: \emph{revalidate} (the default; re-checks every existing specification's statistics on the slid in-sample window with zero new LLM calls) and \emph{repropose} (a full fresh discovery run---new Planner seeds, new proposals, a re-run gate---in the event's own subfolder). \S\ref{sec:wf}'s cross-system rolling comparison uses repropose throughout, since revalidate alone cannot add genuinely new candidates and would not test whether re-discovery itself has value. The in-sample window is a \emph{fixed-length sliding} window, not an expanding one: at re-discovery event with out-of-sample month $m$, the in-sample end date is the month-end immediately preceding $m$, and the in-sample start date is exactly \texttt{window\_length\_years} (10 years) earlier, so a decision at any point in the walk-forward never conditions on more history than a fixed-length lookback allows, however far the walk-forward has progressed. Repropose and combinator-refit cadences are independent config parameters (we evaluate an annual repropose cadence with a 4-month refit cadence, \S\ref{sec:setup}); between two repropose events, the admitted library is held fixed and only the combinator is refit. Two flags independently control what a repropose event inherits from the immediately preceding one: whether the natural-language lesson log carries forward, and whether the admitted-factor library and its registry do. A separate mechanism, used only to isolate the effect of re-discovery itself from ordinary run-to-run LLM variance, lets a rolling arm's very first re-discovery event clone an already-completed static run's initial library wholesale rather than independently rediscovering it, so the two arms' trajectories are identical until the rolling arm's own first library change.

\section{Per-Baseline Fidelity Notes}
\label{app:baselines}
We disclose porting fidelity per system rather than assert uniform faithfulness. Table~\ref{tab:main}'s Productivity metric is each baseline's own admission count---native-gate, except where a baseline's own evaluator has no look-ahead check of its own (RD-Agent(Q)), in which case it is the firewall-clean count instead.

\paragraph{AlphaAgent \citep{alphaagent2025}} proposes a market hypothesis and a factor expression together, constrained at generation time by a complexity cap on the formula's size, then admitted on two independent criteria (largest-common-subtree novelty against the admitted zoo, and an LLM-judged hypothesis-formula consistency score) with no separate backtest-significance gate, matching the paper's stated objective; its continuous regularization weights are never disclosed in the paper, so our port uses documented, reasonable substitutes rather than a literal reconstruction.

\paragraph{AlphaForge \citep{alphaforge2025}} is verified against the authors' own released repository; its generative-predictive mining stage (a generator trained against a frozen predictor, not an adversarial/GAN setup) is replaced with random search over the same, math-verified operator grammar (consistent with the paper's own random-initialization bootstrap, Algorithm 1), while its dynamic-combination stage keeps the paper's real filter-then-regress structure: a rolling-window rank-IC $>$0.02 and rank-ICIR $>$0.2 threshold, falling back to the single best factor if none clears it, before the regression step. It makes zero LLM calls.

\paragraph{RD-Agent(Q) \citep{rdagentq2025}} is verified against the actual installed official package; its evaluator has no temporal-leakage check of its own, only execution success and near-duplicate detection, so our port runs our look-ahead firewall as a post-hoc audit rather than a pre-execution gate, to avoid misrepresenting what the original system actually checks, and reports native, firewall-audited, and firewall-clean views separately: in one matched-budget (300-candidate) run, its native evaluator admitted 57 candidates, of which the firewall subsequently flagged 18 (31.6\%) for temporal look-ahead or cross-stock rank pooling, leaving 39 firewall-clean---the view Table~\ref{tab:main} reports.

\paragraph{QuantaAlpha \citep{quantaalpha2026}} is verified against its real released repository; its complexity/redundancy thresholds are taken directly from its source, a higher-fidelity tier than AlphaAgent's necessarily-substituted ones, though its paper's \emph{stated} pool-admission rule is verified to not match what its own released code implements---we port the paper's stated rule rather than the code's weaker one, so QuantaAlpha remains a distinct comparison point in our results.

\paragraph{Beyond Prompting \citep{beyondprompting2026}} is a close reconstruction from its paper's Algorithm~1, since the paper releases no official code; its gate thresholds, including the Harvey--Liu--Zhu $t{>}3.0$ hurdle it imports \citep{harveyliuzhu2016}, are kept as published even where they differ from our own gate's conventions, to stay faithful to what the paper actually specifies rather than to our own house style.

\section{Formula-Complexity Case Study: Interactions vs.\ Additive Spreads}
\label{app:complexity-case}
A single-run comparison surfaced a structural difference in \emph{what kind} of formula each system tends to admit, independent of any performance number: SEADS's admitted factors are overwhelmingly multiplicative interactions or ratios of two-to-three ranked characteristics, while RD-Agent(Q) and Beyond Prompting admit mostly flat additive/subtractive spreads of exactly two raw characteristics. Table~\ref{tab:complexity-case} quantifies this over admitted factors pooled across SEADS's locked-gate flagship replicates (5 runs each, Panel A and Panel B, deduplicated by name), against RD-Agent(Q)'s and Beyond Prompting's own natively-admitted and promoted factors from matched-protocol comparison runs (\S\ref{sec:main}). Classification is by static analysis of each admitted formula's Python: the top-level operator combining the final \texttt{factor} assignment, tracing through intermediate variable assignments and stripping postfix method chains (\texttt{.astype(...)}, \texttt{.fillna(...)}) to reach the underlying expression, not a manual read.

\begin{table}[h]
\centering
\small
\tabcolsep=3pt
\begin{tabular}{@{}lrrr@{}}
\toprule
Structural type & SEADS & RD-Agent(Q) & Beyond Prompting \\
\midrule
Multiplicative / gate & 83.8\% & 21.1\% & 14.5\% \\
Ratio & 6.3\% & 11.8\% & 3.4\% \\
Additive / subtractive & 9.0\% & 50.0\% & 78.7\% \\
Mixed / other & 0.9\% & 17.1\% & 3.4\% \\
\bottomrule
\end{tabular}
\caption{Structural type of every admitted factor, by system, static-analysis classification of the final formula assignment (Appendix~\ref{app:complexity-case}). SEADS's percentages pool its locked-gate flagship replicates (5 runs each, Panel A and Panel B), deduplicated by name.}
\label{tab:complexity-case}
\end{table}

\paragraph{Representative formulas.} SEADS examples below are real admitted factors, reduced to their underlying expression (stripping the \texttt{groupby('date')...rank(pct=True)} scaffolding each is actually stored and executed as, per the tracing procedure above); RD-Agent(Q) and Beyond Prompting store formulas natively as expressions, so theirs are shown as-is.
\begingroup\raggedright
\begin{itemize}
\item \textbf{SEADS} (cheapness conditioned on low accruals): \texttt{rank(sale\_bev)*(1-rank(taccruals\_at))}
\item \textbf{SEADS} (long momentum conditioned on stable turnover): \texttt{rank(ret\_48\_12)*(1-}\allowbreak\texttt{rank(turnover\_var\_126d))}
\item \textbf{SEADS} (three-variable example): \texttt{rank(sale\_bev)*(1-rank(spi\_at))*(1-}\allowbreak\texttt{rank(ret\_1\_0))}
\item \textbf{RD-Agent(Q)}: \texttt{rank(cash\_at) - rank(debt\_at)}; \texttt{gp\_gr1a - gp\_gr3a} (unranked).
\item \textbf{Beyond Prompting}: \texttt{rank(gp\_at) - rank(at\_gr1)}; \texttt{rank(fcf\_me) - rank(capx\_at)}.
\end{itemize}
\endgroup
Every SEADS example is a product of ranked legs that must each score well simultaneously; every baseline example assumes its two ingredients contribute independently and linearly.

\paragraph{Interaction terms over flat spreads.} A flat difference of two raw characteristics is, for a linear downstream model, exactly spanned by including both characteristics separately --- discovering it as its own factor adds nothing a linear combinator would not already learn on its own. SEADS's \texttt{TRIVIALITY\_RULE}---a deterministic check applied at every complexity tier the Planner assigns a seed (\S\ref{sec:seads-gen})---is built on this reasoning and rejects plain sums/averages for the same reason; the Proposer's rationale requirement additionally asks the model to state \emph{why combining these variables is stronger than either alone}, which a flat spread does not naturally answer. We are optimizing for a different, arguably harder target than standalone statistical significance: incremental value to a portfolio that already holds the full raw characteristic panel, which is exactly what the novelty gate (\S\ref{sec:seads-gate}) exists to protect.

\paragraph{Empirical illustration.} Beyond Prompting's admits illustrate the failure mode this guards against concretely: many hold one ingredient (a cash-flow-yield characteristic) fixed and vary the second across a long list of unrelated characteristics (\texttt{fcf\_yield\_minus\_capex\_intensity}, \texttt{fcf\_yield\_minus\_asset\_growth}, \texttt{fcf\_yield\_minus\_interest\_burden}, \ldots), producing a set with high within-admission correlation (mean pairwise $|\rho|{=}0.281$, effective independent count $3.46$ of 94 nominal admits by a standard $n/(1+(n-1)\bar\rho)$ diversification measure) and correlation with the raw reference panel itself, not because each is individually invalid but because the family is one idea restated many times. SEADS's admitted sets have lower within-admission correlation across its 5 Panel A flagship replicates (mean pairwise $|\rho|{=}0.167$--$0.229$, effective count $3.51$--$4.49$ of 10--18 nominal admits).

\section{Broader Related-Work Landscape}
\label{app:landscape}
\paragraph{Factor-discovery lineage.} Beyond the five re-implemented systems, earlier and adjacent factor-discovery work includes human--AI interactive mining \citep{alphagpt2023}, reinforcement-learning formula search \citep{alphagen2023}, pre-LLM evolutionary mining \citep{alphaevolve2021}, prompted factor conceptualization \citep{chengtang2024}, and semi-automated feature engineering \citep{gptsignal2024}.

\paragraph{Adjacent LLM miners.} A larger population of LLM-based miners each contributes one specific search or optimization mechanism: chain-based generation \citep{chainofalpha2025} (removed by arXiv administrators for a licensing-rights issue; we note its design but do not treat any of its reported numbers as verifiable), an MCTS miner with frequent-subtree avoidance to resist formulaic homogenization \citep{alphajungle2025}, GFlowNet-based search \citep{alphasage2025}, on-graph evolution \citep{alphaprobe2026}, RL screening \citep{alphar12025}, evolutionary sparse portfolios \citep{efs2025}, and reinforcement fine-tuning \citep{rftalpha2026}. A second group contributes an agentic or strategic mechanism instead: multi-agent debate \citep{factormad2025}, strategy-finding agents \citep{kou2025}, a sandboxed execution framework \citep{hubble2026}, market-logic generation \citep{alphalogics2026}, and agentic finance research more broadly \citep{qrafti2026}.

\paragraph{Memory-based precedents.} Skills-and-experience memory with failure constraints \citep{factorminer2026} and memory-driven hypothesis-to-code discovery \citep{xalpha2026} are the closest precedents to our own memory tiers; \S\ref{sec:seads-state} differentiates on the granularity and pipeline-stage specificity of what feeds back, not on the presence of memory itself.

\paragraph{Text-adjacent precedents.} Some systems use price data with text as an idea seed only \citep{factorengine2026,xalpha2026}, use news text directly \citep{llmfactor2024}, use disclosures to predict firm-level returns rather than register a reusable, named characteristic \citep{text2alpha2025}, or use multi-agent debate for discretionary portfolio construction \citep{alphaagents2025}---distinct from our own two panels, both structured numerical data rather than text.

\paragraph{Evaluation infrastructure and realism.} Evaluation infrastructure exists for formulaic alphas \citep{alphaeval2025} and for trading-decision agents more broadly \citep{investorbench2025}; a survey reporting a median test window of only 1.3 years across agent papers surveyed \citep{llmtradingsurvey2024} motivates the realism argument behind \S\ref{sec:method-backtest}. \citet{hyp2factors2026}, evaluated on cryptocurrency rather than equity markets, is OOS-disciplined on a single static split with one modern backbone---the closest prior instance of \S\ref{sec:method-standard}'s standard we are aware of, absent the system-level backtesting of \S\ref{sec:method-backtest}.

\paragraph{Broader agentic-AI context.} General-purpose AI-scientist engines \citep{aiscientist2024,aiscientistv2_2025,alphaevolve2025deepmind} and the broader LLM-agents-in-finance survey literature \citep{llmagentsurvey2024} situate AEAP within the wider landscape of autonomous research agents. Within AEAP but outside factor discovery, \citet{koijenlevy2026,chengchin2024aapm} pursue agentic structural asset pricing and \citet{tradingagents2024,sharp2026,multiagentport2026,finmem2023} pursue portfolio strategy, with \citet{clqt2026} as complementary evaluation infrastructure rather than a portfolio-strategy system itself.

\section{Reproducibility Notes}
\label{app:repro}
\paragraph{Leakage prevention and determinism.} Headline results are drawn only from out-of-sample windows never touched by discovery or selection; the out-of-sample split is excluded from every gate decision in code, and the validation engine is a pure function of a candidate series, the in-sample window, and a threshold dictionary that never reads the out-of-sample split. The firewall, all statistical checks, duplicate/self-collapse fingerprinting, and the sign backstop are deterministic and seed-independent; only the LLM roles are stochastic.

\paragraph{Backbone.} A single current LLM (\S\ref{sec:leakage}) is used for every agentic role in every reported configuration. The engine separately supports resolving a release-dated backbone per historical decision date for a chronologically-consistent alternative protocol; no result in this paper uses that path, consistent with \S\ref{sec:leakage}'s scoping decision.

\paragraph{Shared scoring and replicate reporting.} All systems are scored under one shared portfolio-construction and metric implementation so that ``Sharpe,'' ``admission rate,'' and ``OOS survival'' are defined identically for every system; SEADS and ablation numbers are reported as means $\pm$ standard error over independent replicates (Table~\ref{tab:main}, Table~\ref{tab:ablation-main}), not single runs.

\paragraph{Engineering notes.} Cost asymmetry motivates gate ordering (rank-IC rejects most candidates in milliseconds; partial-IC and Fama--MacBeth each run a per-month regression loop and are placed after the cheaper novelty lookup); the discovery loop supports resumable, checkpointed runs.

\paragraph{LLM configuration.} Every role calls GPT-5.2 through the standard chat-completions API; no role has tool use, function calling, or web access exposed to it. Temperature is set per role rather than fixed globally, matching each role's own need for diversity versus consistency (Table~\ref{tab:llmconfig}).
\begin{table}[t]
\centering
\small
\tabcolsep=4pt
\begin{tabular}{@{}L{2.7cm}cL{2.6cm}@{}}
\toprule
Role & Temp.\ & Why \\
\midrule
Planner, Proposer & 0.7 & breadth across seeds/hypotheses \\
Self-evolution & 0.5 & moderate variation on mutations \\
Implementation Agent, reflector & 0.2 & reliable code/summaries \\
Alignment Checker & 0.1 & consistent judgments over diverse ones \\
\bottomrule
\end{tabular}
\caption{Per-role LLM temperature; all other API parameters are identical across roles.}
\label{tab:llmconfig}
\end{table}

\paragraph{Code release.} All source code required to conduct and analyze the experiments in this paper will be made publicly available upon acceptance, under a license permitting free use for research purposes.

\paragraph{Computing environment.} Python 3.14, pandas 2.2, numpy 2.4, scikit-learn 1.9, scipy 1.17, on macOS; no GPU is used, since all LLM roles call a hosted API rather than running locally, and local computation is limited to lightweight data processing and statistics.

\section{Metric Definitions}
\label{app:metrics}
\paragraph{Rank-IC / ICIR.} Per month, the cross-sectional Spearman correlation $\rho_m$ between factor and next-month excess return over stocks with $\ge5$ valid pairs (months with $<12$ valid $\rho_m$ are undefined, not zero); $\overline{\rho}=\frac1M\sum_m\rho_m$, $\mathrm{ICIR}=\overline{\rho}/\mathrm{std}(\rho_m)\cdot\sqrt{12}$, and the rank-IC $t$-statistic $t=\overline{\rho}/(\mathrm{std}(\rho_m)/\sqrt{M})$. The locked gate thresholds on $t$ alone, floored at 3.0 and rising above that floor via a live Bonferroni bound $t=\Phi^{-1}(1-0.025/N)$ ($\alpha=0.05$, family-wise error rate control) computed from the cumulative candidate count $N$ so far---the only statistic in the gate this correction is applied to, since the Fama--MacBeth bar below is a fixed $t\ge1.0$ regardless of $N$. Coverage is the fraction of in-sample months with a non-null candidate value ($\ge$Min\_Coverage); regime consistency recomputes mean rank-IC on the Regime\_Lookback\_Years immediately preceding the in-sample window and requires sign agreement with it.
\paragraph{Fama--MacBeth \citep{famamacbeth1973} $t$.} Per month, $r_i=a+\sum_k b_k c_{k,i}+b_{\mathrm{cand}}x_i+\varepsilon_i$ over the top-5 most-correlated controls plus the candidate (winsorized 1/99\%, $\ge\max(30,3\cdot7)=30$ stocks/month); $t=\mathrm{mean}(b_{\mathrm{cand}})/(\mathrm{std}(b_{\mathrm{cand}})/\sqrt{T})$.
\paragraph{Partial-IC.} The candidate is residualized each month on its top-5 most-correlated admitted-library columns via per-month OLS; the residual's monthly-Spearman rank-IC $t$-statistic is gated. This top-5 set is computed once from a single shared correlation matrix and reused for novelty, partial-IC, and Fama--MacBeth.
\paragraph{Novelty, duplicate, and self-collapse.} Novelty gates on mean-monthly Spearman $<0.7$ against a designated reference panel plus the admitted library. Duplicate detection uses an exact fingerprint---(count, mean, std, 10th/90th percentile, a hash of the full rounded value array)---so it catches statistically identical candidates without relying on summary statistics alone, which rank-normalized series can share regardless of the underlying variable. Self-collapse flags a final factor correlating $\ge0.95$ with a plain rank of one of its own declared inputs.
\paragraph{Sharpe and its SE.} Monthly decile long-short returns annualized by $\sqrt{12}$; reported gross of costs, with the Lo-(2002) standard error $\mathrm{SE}(\widehat{SR})\approx\sqrt{(1+\widehat{SR}^2/2)/n}$ \citep{lo2002} and a 95\% interval (Table~\ref{tab:se} tabulates this by window length). OOS survival is the fraction of admitted factors with OOS $|t|\ge1.96$ and the same sign as their IS $t$. As a separate diagnostic, not a headline adjustment, we also compute a cost-adjusted return assuming 50bps per side, round-trip, scaled by each factor's own realized monthly rank turnover.
\begin{table}[t]
\centering
\small
\tabcolsep=5pt
\begin{tabular}{@{}lcccc@{}}
\toprule
$n$ (months) & $\widehat{SR}{=}0.3$ & $0.5$ & $0.7$ & $1.0$ \\
\midrule
23 & 0.213 & 0.221 & 0.233 & 0.255 \\
35 & 0.173 & 0.179 & 0.189 & 0.207 \\
59 & 0.133 & 0.138 & 0.145 & 0.159 \\
71 & 0.121 & 0.126 & 0.132 & 0.145 \\
120 & 0.093 & 0.097 & 0.102 & 0.112 \\
\bottomrule
\end{tabular}
\caption{Analytic Lo-(2002) $\mathrm{SE}(\widehat{SR})$ by window length, including Panel B's $n{=}59$.}
\label{tab:se}
\end{table}

\paragraph{Validation thresholds.} Table~\ref{tab:gate} (Appendix~\ref{app:gatefirewall}) gives the gate values used for every result in this paper, alongside each check's execution order: coverage $\ge0.5$, Rank-IC $t\ge\max(3.0,\text{Bonferroni-dynamic})$, out-of-decade regime sign-agreement, novelty $<0.7$, partial-IC $t\ge1.0$, and Fama--MacBeth $t\ge1.0$.

\paragraph{Table~\ref{tab:main}'s cross-replicate standard errors.} Table~\ref{tab:main-se} gives the per-entry standard error (SD across 5 replicates, divided by $\sqrt{5}$) for each of Table~\ref{tab:main}'s four metrics, same systems and panels.
\begin{table}[t]
\centering
\small
\tabcolsep=2pt
\renewcommand{\arraystretch}{0.94}
\begin{tabular}{@{}L{2.9cm}cccc@{}}
\toprule
System & Productivity & Performance & Novelty & Spanning \\
 & (A\,/\,B) & (A\,/\,B) & (A\,/\,B) & (A\,/\,B) \\
\midrule
AlphaForge & 0.45/0.20 & .036/.033 & .025/.028 & .325/.342 \\
RD-Agent(Q) & 2.55/2.81 & .025/.028 & .017/.024 & .364/.295 \\
AlphaAgent & 2.66/2.54 & .017/.023 & .021/.025 & .260/.271 \\
QuantaAlpha & 3.38/2.35 & .022/.027 & .015/.019 & .290/.313 \\
Beyond Prompting & 3.91/3.30 & .017/.014 & .023/.025 & .343/.368 \\
SEADS (ours) & 1.43/1.26 & .025/.021 & .014/.017 & .257/.317 \\
\bottomrule
\end{tabular}
\caption{Per-entry cross-replicate standard errors for Table~\ref{tab:main}, same systems, metrics, and panels.}
\label{tab:main-se}
\end{table}

\section{Example Admits}
\label{app:examples}
Three admitted SEADS factors from the Panel A locked-gate flagship replicates (\S\ref{sec:main}), each chosen to illustrate one property: strong in-sample/out-of-sample agreement, low correlation against the reference panel, and the multiplicative-interaction structure typical of SEADS's admits (Appendix~\ref{app:complexity-case}). All three cleared every check in Table~\ref{tab:gate} (Appendix~\ref{app:gatefirewall}); IS stats are from the in-sample window, OOS from the held-out window (\S\ref{sec:setup}), both from Table~\ref{tab:examples}. These three are hand-selected to each demonstrate one property, not randomly drawn or representative of the admitted pool's average statistics---Table~\ref{tab:main}'s Panel A mean per-factor OOS Sharpe is 0.25, well below any shown here.

\begin{table}[t]
\centering
\small
\tabcolsep=4pt
\begin{tabular}{@{}lccc@{}}
\toprule
 & Strong IS/OOS & Low corr.\ & Represent.\ \\
\midrule
Coverage & 97.8\% & 89.3\% & 85.0\% \\
Rank-IC (ICIR) & 0.024 (1.25) & 0.031 (1.52) & 0.016 (1.35) \\
Partial-IC $t$ & 2.45 & 4.77 & 3.25 \\
FM $t$ & 2.65 & 3.36 & 2.61 \\
OOS Sharpe & 0.89 & 0.52 & 0.88 \\
OOS $t$ & 2.16 & 1.27 & 2.14 \\
OOS ann.\ ret.\ & 13.1\% & 7.1\% & 7.3\% \\
Max corr & 0.662 & 0.449 & 0.697 \\
\bottomrule
\end{tabular}
\caption{Gate and out-of-sample statistics for the three example admits below, one column each; Coverage, Rank-IC (ICIR), Partial-IC $t$, FM $t$, and Max corr are in-sample, computed at admission time, while the three OOS-labeled rows are out-of-sample.}
\label{tab:examples}
\end{table}

\begingroup\raggedright
\paragraph{Strong IS/OOS agreement: \texttt{stable\_liquidity\_}\allowbreak\texttt{efficiency\_gate}} (replicate a, rich tier). \emph{Formula:} \texttt{(1-rank(bidaskhl\_21d))} $\times$ \texttt{(1-rank(zero\_trades\_126d))} $\times$ \texttt{(1-rank(dolvol\_var\_126d/(1+trail12m\_mean)))}. Tight spreads and low trading inactivity indicate low trading frictions and faster price discovery, lowering required returns; downweighting names with an unstable dollar-volume regime isolates the structural, not episodic, component. Correlates most with \texttt{dolvol\_var\_126d} (Table~\ref{tab:examples}).

\paragraph{Low correlation: \texttt{distress\_amplified\_}\allowbreak\texttt{perf\_mispricing}} (replicate d, simple tier). \emph{Formula:} \texttt{rank(mispricing\_perf)} $\times$ \texttt{rank(o\_score)}. Performance-based mispricing is more likely sustained by limits-to-arbitrage and forced trading among financially weak (high O-Score) firms, so the subsequent correction should be stronger when both signals coincide. Correlates most with \texttt{debt\_at}, at 0.449 the lowest of any admit pooled across the 5 replicates (Table~\ref{tab:examples}).

\paragraph{Representative structure: \texttt{cash\_op\_profit\_x\_low\_}\allowbreak\texttt{tax\_payable\_growth}} (replicate b, simple tier). \emph{Formula:} \texttt{rank(cop\_bev)} $\times$ \texttt{(1-rank(txp\_gr1a))}. Cash operating profitability is more informative when low growth in taxes payable reflects a persistent cash-tax advantage rather than a temporary accounting effect. A two-variable rank product, the structural form 83.8\% of SEADS's admits share (Appendix~\ref{app:complexity-case}); correlates most with \texttt{txp\_gr1a} (Table~\ref{tab:examples}).
\endgroup

\section{AI Acknowledgement}
\label{app:ai-ack}
Language models are this paper's experimental subject, not merely a tool used alongside it: every reported result comes from an LLM-driven system, SEADS and all five baselines alike, which call a large language model at every proposal, evaluation, and evolution step (\S\ref{sec:aap}, Appendix~\ref{app:internals}). Separately, generative AI tools (large language models) were also used as an assistant in this project's own process---for part of the supporting code development, and for portions of this manuscript's preparation and editing. All experimental results, data, and claims were verified by the authors against the underlying code and run logs; the authors take full responsibility for the paper's content.

\end{document}